\documentclass[11pt,a4paper]{article}
\usepackage{times,latexsym}
\usepackage{url}
\usepackage{soul}
\usepackage{authblk}
\usepackage[T1]{fontenc}

\usepackage{amssymb}

  \usepackage[acceptedWithA]{tacl2018v2}
\usepackage[]{tacl2018v2}

\usepackage{xspace,mfirstuc,tabulary}

\usepackage{graphicx}

\usepackage{array, multirow}
\usepackage{tabularx}
\usepackage{pbox}
\usepackage{amsmath}
\usepackage{amssymb}

\usepackage{makecell}
\usepackage{textpos}
\usepackage{xcolor}
\usepackage{arydshln}

\newif\iftaclinstructions
\taclinstructionsfalse 
\iftaclinstructions
\renewcommand{\confidential}{}
\renewcommand{\anonsubtext}{(No author info supplied here, for consistency with
 TACL-submission anonymization requirements)}
\newcommand{\instr}
\fi

\iftaclpubformat 

\else

\fi

\DeclareMathOperator*{\argmax}{argmax}

\title{Combining Pre-trained Word Embeddings and Linguistic Features\\ for Sequential Metaphor Identification}

\author[1]{\textbf{Rui Mao}}
\author[2]{\textbf{Chenghua Lin}}
\author[1]{\textbf{Frank Guerin}}
\affil[1]{Department of Computing Science, University of Aberdeen, AB24 3UE, UK} 
\affil[1]{\vspace*{0.1cm}\tt {\{r03rm16, f.guerin\}@abdn.ac.uk}}
\affil[2]{Department of Computer Science, University of Sheffield, S1 4DP, UK}
\affil[2]{\tt {c.lin@sheffield.ac.uk}}
\date{}

\begin{document}
\maketitle
\begin{abstract}
We tackle the problem of identifying metaphors in text, treated as a sequence tagging task.
The pre-trained word embeddings GloVe, ELMo and BERT have individually shown good performance on sequential metaphor identification.
These embeddings are generated by different models, training targets and corpora, thus encoding different semantic and syntactic information. 
We show that leveraging GloVe, ELMo and feature-based BERT based on a multi-channel CNN and a Bidirectional LSTM model can significantly outperform any single word embedding method and the combination of the two embeddings. Incorporating linguistic features into our model can further improve model performance, yielding state-of-the-art performance on three public metaphor datasets.
We also provide in-depth analysis on the effectiveness of leveraging multiple word embeddings, including analysing 
the spatial distribution of different embedding methods for metaphors and literals, and showing how well the embeddings complement each other in different genres and parts of speech.
\end{abstract}

\section{Introduction}
\label{sect: introduction}


Linguistically, metaphor is defined as a figurative expression that uses one or several lexical units to represent a different meaning in the context, where there is a semantic contrast between the contextual and basic meanings of the lexical units~\citep{group2007mip}. Metaphors are widely found in corpora, causing issues for natural language processing (NLP) tasks, such as sentiment analysis \citep{ghosh2015semeval} and machine translation \citep{mao2018word}. Traditionally, metaphors are identified from dependent word-pairs, e.g. adjective-noun and verb-noun pairs \citep{turney2011literal, neuman2013metaphor, rei2017grasping}. There is an increasing trend to attempt metaphor identification as an end-to-end sequence tagging task \citep{wu2018neural, gao2018neural, mao2019metaphor}, as the settings of  traditional approaches are not practical in real-world applications. In contrast, a sequential metaphor identification model can predict the metaphoricity of all words in a sentence, as a sequence labelling task, without needing to locate word-pairs beforehand. 




Pre-trained word embeddings have been shown to provide significant improvements to the state of the art for a wide range of NLP tasks~\citep{young2018recent}. Currently, there are many choices of pre-trained word embeddings, such as GLoVe \citep{pennington2014glove}, ELMo \citep{peters2018deep} and BERT \citep{devlin2019bert}. These word embeddings are based on different training targets, training sets,  and model frameworks. E.g., GloVe is trained via token co-occurrence distributions, whose vectors are independent of the context of downstream tasks. ELMo is a BiLSTM based context dependent representation, whose training target is training Language Models (LM) from two directions. BERT is also context dependent, and is trained to predict masked words and next sentences via the Bidirectional Transformer \citep{vaswani2017attention}. All the aforementioned pre-trained word embeddings have presented strong performance on a variety of NLP tasks. Our hypothesis is that different word embeddings may capture different semantics and syntax knowledge, and thus metaphor identification models may benefit from their combination  by taking advantage of their complementarity. 

The motivation of this paper is to investigate the utilities of different pre-trained word embeddings, e.g., GloVe, ELMo, feature-based BERT and their combination (GEB) for metaphor identification. 
In particular, if we can combine existing word embeddings to take advantage of their respective strengths, then we do not have to re-train a supermodel on a multi-genre corpus for every technological breakthrough in word embedding. 
To facilitate our investigation, we propose a neural architecture consisting of a multi-channel CNN layer and a BiLSTM layer, where each CNN channel takes one embedding as input. Combining CNN and BiLSTM also allows the model to capture local and long-term dependent information of an input sequence.

We evaluate on three publicly available datasets, namely VUA \citep{steen2010method}, MOH-X \citep{mohammad2016metaphor} and TroFi \citep{birke2006clustering} datasets. Experiments show that our GEB model yields competitive performance for sequential metaphor identification compared to the strongest baseline. Incorporating linguistic features (i.e., PoS and word abstractness \citep{rabinovich2018learning}) into the GEB model achieves further gains, yielding the state-of-the-art performance for sequential metaphor identification. 
In terms of the effectiveness of different embeddings, using the combination model GEB yields better results than using each embedding individually or the combinations of any two of them. In addition, it was found that GloVe, ELMo and BERT have different strengths in modelling metaphors from different genres and in different parts of speech (PoS), 
because they are trained on different corpora with different algorithms. This explains why the combination outperforms the individuals.

The contribution of this paper can be summarised as follows: (1) we propose a 3-channel CNN based model\footnote{Our code will be released after blind review.} that can incorporate diverse embeddings and features for sequential metaphor identification, achieving state-of-the-art performance; (2) we investigate the utilities of the combinations of GloVe, ELMo and feature based BERT across 24 layers, showing their complementarity in terms of different genres and PoS; (3) we provide insight into the factors determining which combinations of embeddings that are likely to have good performance on a downstream task, e.g., maintaining a balance between general and task-specific features. 






\section{Related Work}

Traditionally, metaphors are identified from dependent word-pairs with various feature engineering methods. Psychological features such as abstractness and concreteness have been widely applied in verb-noun and adjective-noun metaphor identification tasks \citep{turney2011literal, neuman2013metaphor, assaf2013dark, tsvetkov2014metaphor}. Their hypothesis is that metaphors normally are concrete, associated with an abstract dependent word. \citet{shutova2016black, shutova2017multilingual} and \citet{rei2017grasping} classified metaphors by modelling word distribution features with supervised and unsupervised models. \citet{mao2018word} extended the word-pair approaches by identifying metaphors from full sentences, taking the full context of a sentence into account using input and output vectors of word2vec \citep{mikolov2013distributed}. However, these methods are based on a condition that the position of a target word whose metaphoricity to be identified in a sentence has been annotated in advance. Such approaches are inconvenient in real world applications. 

Sequential metaphor identification takes one step further because the metaphoricity of all words in a sentence can be identified without knowing the position of a target word in advance. \citet{wu2018neural} employed a one-channel CNN and BiLSTM based framework for sequential metaphor identification. Their model encodes concatenated word2vec, word2vec cluster labels and PoS labels. Then, the hidden states of BiLSMT are classified by a softmax classifier. This model with an ensemble learning strategy achieved the best performance on the NAACL-2018 Metaphor Shared Task \citep{leong2018report}. \citet{gao2018neural} used a vanilla BiLSTM based sequence tagging model. The concatenation of GloVe and ELMo is encoded by BiLSTM and then classified by a softmax classifier to predict a sequence of metaphoricity labels. \citet{mao2019metaphor} proposed two linguistics informed models. The first Metaphor Identification Procedure (MIP) \citep{group2007mip} inspired model emphasised the contrast between contextual and literal meanings of a target word. The second Selectional Preference Violation (SPV) theory \citep{wilks1975preferential, wilks1978making} inspired model highlighted the contrast between a metaphor and its context. Both models outperform the models of \citet{wu2018neural} and \citet{gao2018neural}, while the SPV informed model is the state-of-the-art.

While \citet{gao2018neural} and \citet{mao2019metaphor} have used more than one word embeddings for metaphor identification, the number of embeddings was limited at two, and more importantly they did not explore how and why combining pre-trained embeddings is effective for the task. We seek to answer this question by providing systematic evaluation on the efficiency of a variety of embedding combinations as well as their complementarity, leading to interesting findings relating to genre and PoS types of the corpus for training word embedding models. 
In addition, linguistic features such as abstractness have only been used in traditional machine learning based verb-noun and adjective-noun pair metaphor identification~\citep{turney2011literal, neuman2013metaphor, tsvetkov2014metaphor}. We  examine the use of abstractness and PoS features for sequential metaphor identification.

\section{Methodology}

We propose a CNN-BiLSTM framework for sequential metaphor identification, partially inspired by the works of \citet{kim2014convolutional} and \citet{wu2018neural}. One of the key differences between our model and that of \citet{wu2018neural} is that we use a multi-channel CNN layer to incorporate different word embedding features, whereas \citeauthor{wu2018neural} simply used a single-channel CNN on word2vec. 
Note that using a multi-channel CNN component allows us to combine different pre-trained word embeddings in a simple way. Additionally, CNN has shown its effectiveness on a variety of sequence tagging tasks \citep{poria2016aspect, chen2017feature, xu_acl2018}, but these works have only explored using a single-channel CNN and a single word embedding. We give the technical details of our model in the following sections.
\subsection{Framework} \label{sec:framwork}

\begin{figure*}[tb]
    \centering
    \includegraphics[width=6.2in]{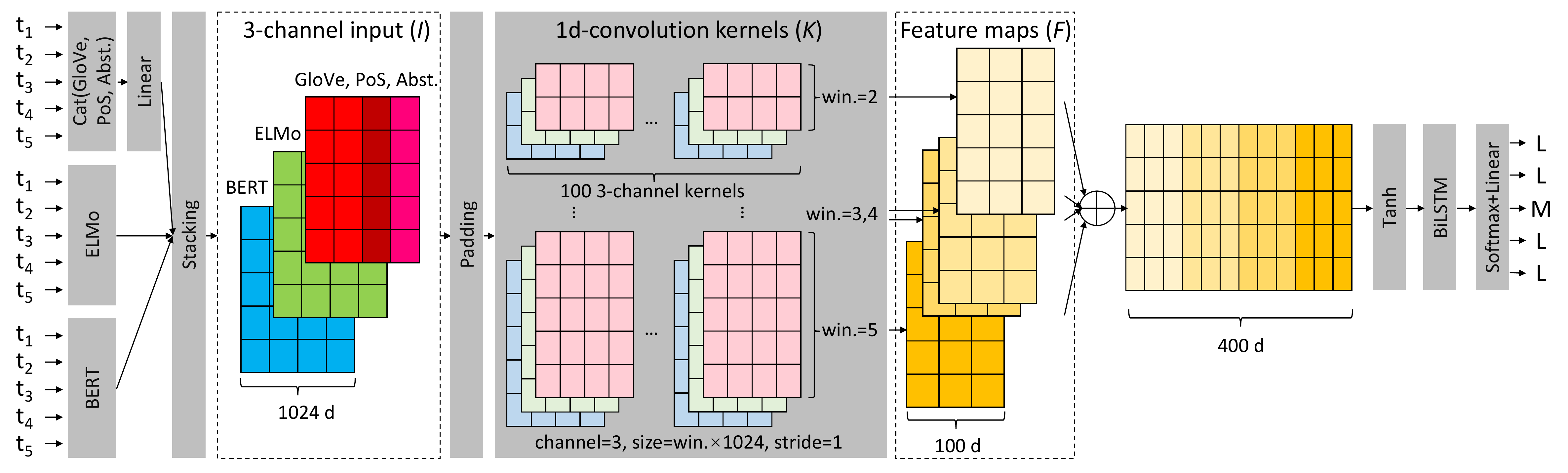}
    \caption{Model framework. Grey boxes are embedding and computational layers. $\oplus$ denotes concatenating. 
    }
    \label{fg: framework}
\end{figure*} 

Figure~\ref{fg: framework} shows the overall architecture of our model. We apply a CNN operation on the input which has three channels for leveraging different embeddings. The pre-trained contextual embeddings ELMo and BERT have the same dimensionality, which is 1024. However, GloVe only has 300 dimension, so we project GloVe into 1024 dimensions with a linear layer to align the dimensionality of all the input embeddings.
\begin{equation}
\label{eq:linear1}
    \mathbf{M}_G=W^{\phi} \cdot \text{GloVe} + b^{\phi}
\end{equation}
Here $W^{\phi}$ is the parameter matrix of the linear layer, and $b^{\phi}$ is its bias vector.
We then stack the input embedding matrices (i.e., the projected GloVe, ELMo and BERT) which is used as input  for the subsequent feature map calculation. 
\begin{equation} \label{eq:input1}
\mathbf{I} = 
\begin{bmatrix}
\mathbf{M}_G & \mathbf{M}_E  & \mathbf{M}_B
\end{bmatrix}
\end{equation}
Note that we pad zero vectors for $\mathbf{I}$ to make the row dimensionality  of the feature map matrix the same as the length of the input sequence.

An alternative version of our model  also incorporates external linguistic features for model learning. In our study, we experimented with PoS and word abstractness ($Abst$) features, which are represented in one-hot encoding and real numbers ($\in [0,1]$), respectively. We concatenate ($\oplus$) the linguistic feature representations with Glove
\begin{equation}
\label{eq:input2}
\mathbf{M}_{GPA} = \text{GloVe} \oplus \text{PoS} \oplus \text{Abst}.
\end{equation}
$\mathbf{M}_{GPA}$ is then projected to 1024 dimension via a linear layer (Eq.~\ref{eq:linear1}) to make an alternative version of $\mathbf{M}_G$. Finally, the resulting representation is combined with ELMo and BERT in a similar way (Eq.~\ref{eq:input1}) as previously described. 

Given a 3-channel embedding input ($\mathbf{I}$), a feature map ($F$) is calculated as follows: 
\begin{equation}
\label{eq:feature_map}
    F_{i, w} = \sum_{c} \sum_{w} \sum_{d} \mathbf{I}_{(c, i+w-1, d)}K_{(c,w,d)}, \\
\end{equation}   
where $F_{i,w}$ is the new representation of the $i$th word of an input sentence obtained by applying CNN operations on input embeddings for the corresponding word with kernel $K$ and window size $w$. $c,d$ are channel size ($c =3$) and the dimensionality of input word embeddings ($d =1024$), respectively.  In order to capture different local contextual information, we use CNN kernels with four types of window sizes $w =\{2,3,4,5\}$ and for each window size 100 kernels.  This results in four different feature maps, each of which corresponds to a particular window size. 
Finally, we concatenate all the feature maps together 
\begin{equation}
    F = [F_{w=2} \oplus F_{w=3} \oplus F_{w=4} \oplus F_{w=5}]. 
\end{equation}



We fed $F$ with a Tanh activation function into a BiLSTM layer as input. A BiLSTM is employed for capturing long-term dependencies of $F$, because the CNN layer only encodes the contextual information of a word in $n$-grams
\begin{equation}
    h_i = \text{BiLSTM}(\text{Tanh}(F_i), \overrightarrow{\strut h}_{i-1}, \overleftarrow{\strut h}_{i+1}),
\end{equation}
where $h$ is a hidden state of BiLSTM. Finally, the metaphoricity ($y_i$) for a given word is predicted by a softmax ($\sigma$) classifier
\begin{equation}
    p(y_i) = \sigma (W_i^{\psi} \cdot h_i+b_i^{\psi}), 
\end{equation}
where $W^{\psi}$ and $b^{\psi}$ are model parameters and biases estimated by optimising the following objective function based on weighted cross entropy loss.
\begin{equation}
    \mathcal{L} = - \sum_{i} \omega_{y_i} y_i \log (\hat{y}_i)
\end{equation}
Following \citet{wu2018neural} and \citet{mao2019metaphor}, we set the weight ($\omega$) for metaphors and literals to 2 and 1, respectively. 
\subsection{Features}
We select GloVe, ELMo and BERT because they were trained on different corpora with different genres and training targets, which may exhibit distinctive semantic and syntactic characteristics. 
For instance, GloVe was trained on Common Crawl\footnote{\url{http://commoncrawl.org}}, from billions of web pages (840 billion tokens); ELMo was trained on WMT 2011 News Crawl data\footnote{\url{http://www.statmt.org/lm-benchmark/}} (800 million tokens); BERT was trained on Wikipedia\footnote{\url{https://dumps.wikimedia.org/}} (2.5 billion tokens) and BookCorpus\footnote{\url{https://yknzhu.wixsite.com/mbweb}} (800 million tokens) \citep{zhu2015aligning}, where scientific articles and novels are included. 
We employed feature-based BERT rather than fine-tuned BERT because the former allows us to investigate the utility of each BERT layer of a fixed depth neural network.
We hypothesise that sequential metaphor identification will benefit from the complementarity of multiple word embeddings.



Apart from embedding features, we also explored linguistic features, i.e., PoS and abstractness of words, in order to examine whether these universal features can further improve our model performance.  
The abstractness feature set ($\mathcal{V}$) contains 99,954 words  with each word assigned with an abstractness score ranging from 0 to 1~\cite{rabinovich2018learning}, i.e., the higher the score, the more abstract the corresponding word. 
For instance, the abstractness score of \textit{purism} is 0.97 (abstract), while abstractness of \textit{ski} is 0.25 (concrete). For a word $w_{o}$ which is not listed in $\mathcal{V}$, we predict its abstractness by taking the abstractness score of its most semantically similar word ($w_s$) in $\mathcal{V}$, where the calculation is based on GloVe ($g$) using cosine similarity. 
\begin{align}
\begin{gathered}
w_{s} = \argmax_{w_i \in \mathcal{V}}  \text{cosine}(g_{w_{o}}, g_{w_{i}}) \\
Abst_{w_{o}} = Abst_{w_{s}}
\end{gathered}
\end{align}
In  case $w_o$ is out of the vocabulary of GloVe, it will be assigned with an abstractness score of 0.5. 

\section{Experiment}
\label{sect: experiment}

\begin{table}[t!]
\small
\begin{center}
\begin{tabular}{l|r >{\centering\arraybackslash}p{0.8cm} >{\centering\arraybackslash}p{0.8cm}  >{\centering\arraybackslash}p{0.8cm}}
\Xhline{2.5\arrayrulewidth}
 \bf Dataset & \pbox{21cm}{\bf \# Tgt \\ token} & \pbox{21cm}{\bf \% \\ M} & \bf \# seq
 & \pbox{21cm}{\bf Avg \#\\ M/S}\\
\hline 
VUA\_all & 205,425 & 11.6 & 10,567 & 3.4\\
VUA\_trn & 116,622 & 11.2 & 6,323 & 3.3\\
VUA\_dev & 38,628 & 11.6 & 1,550 & 4.0\\
VUA\_tst & 50,175 & 12.4 & 2,694 & 3.4\\
VERB\_tst & 5,873 & 30.0 & 2,694 & 1.5 \\
\hline 
MOH-X & 647 & 48.7 & 647 & 1.0\\
\hline
TroFi & 3,737 & 57.6 & 3,737 & 1.0\\
\Xhline{2.5\arrayrulewidth}
\end{tabular}
\end{center}
\caption{\label{tb: dataset} Dataset statistics \citep{mao2019metaphor}. NB: \# Tgt token is the number of target tokens whose metaphoricity is to be identified. \% M is the percentage of metaphoric tokens among target tokens. \# seq is the number of sequences. 
Avg \# M/S is the average number of metaphors per metaphorical sentence.}
\end{table}
\begin{table*}[t]
\small
\centering
\begin{tabular}{>{\centering\arraybackslash}p{2.5cm}|>{\centering\arraybackslash}p{0.36cm}>{\centering\arraybackslash}p{0.36cm}>{\centering\arraybackslash}p{0.36cm}>{\centering\arraybackslash}p{0.41cm}|>{\centering\arraybackslash}p{0.36cm}>{\centering\arraybackslash}p{0.36cm}>{\centering\arraybackslash}p{0.36cm}>{\centering\arraybackslash}p{0.41cm}|>{\centering\arraybackslash}p{0.36cm}>{\centering\arraybackslash}p{0.36cm}>{\centering\arraybackslash}p{0.36cm}>{\centering\arraybackslash}p{0.41cm}|>{\centering\arraybackslash}p{0.36cm}>{\centering\arraybackslash}p{0.36cm}>{\centering\arraybackslash}p{0.36cm}>{\centering\arraybackslash}p{0.41cm}}
\Xhline{2.5\arrayrulewidth}
\multirow{2}{*}{\bf Model} & \multicolumn{4}{c|}{\bf VUA ALL POS} & \multicolumn{4}{c|}{\bf VUA VERB} & \multicolumn{4}{c|}{\bf MOH-X (10-fold)} & \multicolumn{4}{c}{\bf TroFi (10-fold)} \\
& \bf P & \bf R & \bf F1 & \bf Acc & \bf P & \bf R & \bf F1 & \bf Acc & \bf P & \bf R & \bf F1 & \bf Acc & \bf P & \bf R & \bf F1 & \bf Acc\\
\hline
\citeauthor{wu2018neural} & 60.8 & 70.0 & 65.1 & - & 60.0 & 76.3 & 67.2 & - & 69.2 & 69.9 & 69.6 & 70.0 & 79.6 & 78.8 & 79.2 & 78.3 \\
\citeauthor{gao2018neural} & 71.6 & 73.6 & 72.6 & 93.1 & 68.2 & 71.3 & 69.7 & 81.4 & 79.1 & 73.5 & 75.6 & 77.2 & 87.7 & 87.4 & 87.6  & 86.9\\
\citeauthor{mao2019metaphor}-SPV & 73.0 & 75.7 & \underline{74.3} & 93.8 & 66.3 & 75.2 & \underline{70.5} & 81.8 & 77.5 & 83.1 & \underline{80.0} & 79.8 & 89.8 & 88.1 & \underline{88.9} & 88.0\\
\hline
GEB$_{17}$ & 74.9 & 74.4 & 74.7* & 93.9 & 70.4 & 72.1 & 71.2* & 82.5 & 78.0 & 83.1 & 80.4* & 80.2 & 90.7 & 89.0 & 89.8* & 88.4\\
PoS+Abst+GEB$_{17}$ & 72.5 & 77.4 & \bf 74.9* & 94.0 & 68.8 & 74.5 & \bf 71.5* & 82.4 & 77.9 & 83.8 & \bf 80.7* & 80.3 & 89.3 & 91.0 & \bf 90.2* & 88.6\\
\Xhline{2.5\arrayrulewidth}
\end{tabular}
\caption{Model performance. * denotes $p < 0.01$ on a two-tailed t-test, against the best baseline with an underline.}
\label{tb: results}
\end{table*}

\subsection{Dataset}
We examine our model on three public datasets, whose statistics are summarised in Table~\ref{tb: dataset}.

\noindent \textbf{VUA}~~~VU Amsterdam Metaphor Corpus \citep{steen2010method} is the largest all-word annotated metaphor dataset (Fleiss' Kappa = 0.84), whose sentences are sampled from the British National Corpus \citep{leech1992100}. The dataset contains four genres (i.e., academic, conversation, fiction, and news) and has been used in the NAACL-2018 Metaphor Shared Task \citep{leong2018report}. Following the shared task, we construct two versions of the datasets, i.e., VUA-all-PoS and VUA-verb. VUA-all-PoS considers all words as target words, whereas VUA-verb takes a subset of verbs in VUA as target words. 
VUA is our main dataset for the later breakdown analysis ($\S$~\ref{sect: compensations analysis}) as it contains metaphors from different genres and PoS.

\noindent \textbf{MOH-X}~~~The dataset was formed by sampling example sentences from WordNet \citep{fellbaum2005wordnet} by \citet{mohammad2016metaphor}. A named target verb for each sentence was annotated by 10 participants. In MOH-X, each metaphoricity label has been agreed by at least 70\% annotators. There are 453 unique target verbs in MOH-X.

\noindent \textbf{TroFi}~~~The dataset was sampled from the 1987-89 Wall Street Journal Corpus Release 1 \citep{charniak2000bllip}, where target verbs of 3,737 sentences were annotated by \citet{birke2006clustering}. We sample these human annotated sentences of the original TroFi dataset in our tests. There are 50 unique target verbs in TroFi. 

For MOH-X and TroFi, we consider the rest of words (non-targets) as literal for the model training, because only a single target word in a sentence is annotated, which is in line with \citet{gao2018neural}. Our model is evaluated on the target words with 10-fold cross validation on these two datasets. 

\subsection{Baselines}
We compare our model against three strong baselines on the task of sequential metaphor identification. Details of the baselines are given below.

\noindent \textbf{\citet{wu2018neural}} proposed a CNN+BiLSTM model, fitting word2vec, PoS, word2vec cluster features to sequential metaphor identification tasks with an ensemble learning strategy.

\noindent \textbf{\citet{gao2018neural}} adopted a standard sequence tagging model with a BiLSTM encoding layer. They used ELMo and GloVe as input features.

\noindent \textbf{\citet{mao2019metaphor}} proposed two models inspired by linguistic theories (MIP and SPV), claiming
state-of-the-art performance across three bench mark metaphor datasets using ELMo and GloVe. 


\subsection{Setup}
We adopt 840B GloVe\footnote{\url{http://nlp.stanford.edu/data/glove.840B.300d.zip}}, original ELMo\footnote{\url{https://allennlp.org/elmo}} and cased BERT large 
pre-trained word embeddings. PoS tags are parsed by spaCy. 
We adopt four kernel sizes ($2 \times 1024$, $3 \times 1024$, $4 \times 1024$, and $5 \times 1024$) with a stride of 1 in the convolution layer. Concatenated feature maps are activated by the Tanh function. The dimension of BiLSTM hidden states is $256 \times 2$. We use SGD optimizer with a learning rate of 0.2. Dropout is applied on 3-channel input and BiLSTM hidden states with rates of 0.5 and 0.1, respectively.

\section{Results and discussion}
\subsection{Overall performance}
\label{sect: overall performance}

F1 score is our main measure, where metaphor is the positive class. As shown in Table \ref{tb: results}, our model GEB$_{17}$ (the subscript 17 denotes the 17th BERT layer) achieves better performance than the state-of-the-art baselines. By incorporating linguistic features, our model PoS+Abst+GEB$_{17}$ yields extra performance gains, outperforming the best baselines on all datasets significantly. 

When comparing GEB$_{17}$ against \citet{wu2018neural}, the baseline model which has the most similar architecture to ours, we see that using multiple CNN channels and word embeddings, especially contextual dependent word embeddings, can substantially improve CNN-BiLSTM model performance by an average of 8.8\% in F1 across all datasets.
GEB$_{17}$ also outperforms the models of \citet{gao2018neural} and \citet{mao2019metaphor} which both utilise GloVe and ELMo, with an average gain of 2.7\% and 0.6\% in F1, respectively. One may notice that the gain of GEB$_{17}$ over the SPV model \cite{mao2019metaphor} is relatively small. This is likely because SPV benefits from the use of extra multi-head contextual attention mechanism whereas our encoder is simpler with only BiLSTM. 
Overall, the experimental results have clearly demonstrated the effectiveness of employing multiple word embeddings for the task of metaphor identification. 

In another set of experiments, we further explore whether universal linguistic features (e.g., word abstractness and PoS) that are not explicitly encoded by pre-trained word embeddings can further improve deep learning models for sequential metaphor identification. Experimental results show that incorporating word abstractness and PoS features does help boosting model performance but not substantially, where the gain of PoS+Abst+GEB$_{17}$ over GEB$_{17}$ is around 0.3 F1 on average. 
This is probably because contextual dependent word embeddings have already implicitly learnt linguistic features such as PoS from corpora \citep{reif2019visualizing}, resulting in the explicit incorporation of linguistic features being less effective (see $\S$ \ref{sect:POS} for detailed discussion). 

\begin{table}[t]
\small
\begin{center}
\begin{tabular}{c| >{\centering\arraybackslash}p{0.45cm} >{\centering\arraybackslash}p{0.45cm} >{\centering\arraybackslash}p{0.45cm} >{\centering\arraybackslash}p{0.45cm} >{\centering\arraybackslash}p{0.45cm} >{\centering\arraybackslash}p{0.45cm}}
\Xhline{2.5\arrayrulewidth}
  & \bf VB & \bf ADJ & \bf NN & \bf ADV & \bf ALL\\
\hline 
GEB$_{17}$ & 69.4 & 57.8 & 61.3 & 57.6 & 72.5\\
PoS+GEB$_{17}$ & 69.4 & 55.3 & \bf 63.4 & 62.1 & 72.8 \\
Abst+GEB$_{17}$ & 69.1 & 55.2 & 61.7 & 59.8 & 72.3 \\
PoS+Abst+GEB$_{17}$ & \bf 70.7 & \bf 58.2& \bf 63.4 & \bf 62.4 & \bf 73.5 \\
\Xhline{2.5\arrayrulewidth}
\end{tabular}
\end{center}
\caption{\label{tb: posabst} Linguistic feature analysis on VUA-all-PoS development set, measured by F1 score.}
\end{table}

\begin{table*}[t!]
\small
\begin{center}
\begin{tabular}{l|cc|cc|cc|cc|cc}
\Xhline{2.5\arrayrulewidth}
\multirow{2}{*}{} & \multicolumn{2}{c|}{\textbf{Verb}} & \multicolumn{2}{c|}{\textbf{Adjective}} & \multicolumn{2}{c|}{\textbf{Noun}} & \multicolumn{2}{c|}{\textbf{Adverb}} & \multicolumn{2}{c}{\textbf{Overall}} \\
\multirow{2}{*}{} & \textbf{Meta} & \textbf{Lite} & \textbf{Meta} & \textbf{Lite} & \textbf{Meta} & \textbf{Lite} & \textbf{Meta} & \textbf{Lite} & \textbf{Meta} & \textbf{Lite} \\
\hline
Mean              & 0.26           & \textbf{0.28}  & 0.27           & \textbf{0.34}  & 0.33           & \textbf{0.36} & \textbf{0.25}  & 0.22           & 0.29           & \textbf{0.30}  \\
Std.              & 0.18           & 0.16           & 0.19           & 0.22           & 0.21           & 0.21 & 0.21           & 0.18           & 0.21           & 0.19           \\
1st Quart.        & 0.11           & \textbf{0.16}  & 0.13           & \textbf{0.16}  & 0.17           & \textbf{0.19} & \textbf{0.13}  & 0.11           & \textbf{0.15}  & 0.14  \\
Median           & 0.21           & \textbf{0.29}  & 0.22           & \textbf{0.26}  & 0.28           & \textbf{0.33} & \textbf{0.18}  & \textbf{0.18}  & 0.21           & \textbf{0.24}  \\
3rd Quart.        & \textbf{0.32}  & 0.30           & 0.35           & \textbf{0.50}  & 0.48           & \textbf{0.51} & \textbf{0.31}  & 0.23           & \textbf{0.43}  & \textbf{0.43}  \\
\Xhline{2.5\arrayrulewidth}
\end{tabular}
\end{center}
\caption{\label{tb: abst} Abstractness statistics for metaphors and literals.}
\end{table*}

We ran ablation experiments to test the utilities of PoS and abstractness with GEB$_{17}$ on VUA-all-PoS development set. As seen in Table \ref{tb: posabst}, the gain of PoS+GEB$_{17}$ over GEB$_{17}$ is small (0.3 F1 on all PoS). Additionally, simply adding abstractness features (Abst+GEB$_{17}$) does not improve the GEB$_{17}$ model performance. This is because the abstractness level is less distinctive between metaphors and literals overall (see Table~\ref{tb: abst}). Although metaphorical verbs, adjectives and nouns have lower abstractness scores than literals in terms of mean, first quartile and median, adverbs hold the opposite. By combining PoS and abstractness features, PoS+Abst+GEB$_{17}$ yields a gain of 1.0 F1 against GEB$_{17}$ on VUA-all-PoS development set, where the largest improvement appears in adverbs. This is somewhat unexpected as using each of the linguistic features alone is not effective. A plausible explanation is that guided by explicit PoS features, our model can better distinguish and make use of the more useful abstractness features for metaphor identification.
\begin{center}
\begin{figure}[t!]
    \centering
    \includegraphics[scale=0.47]{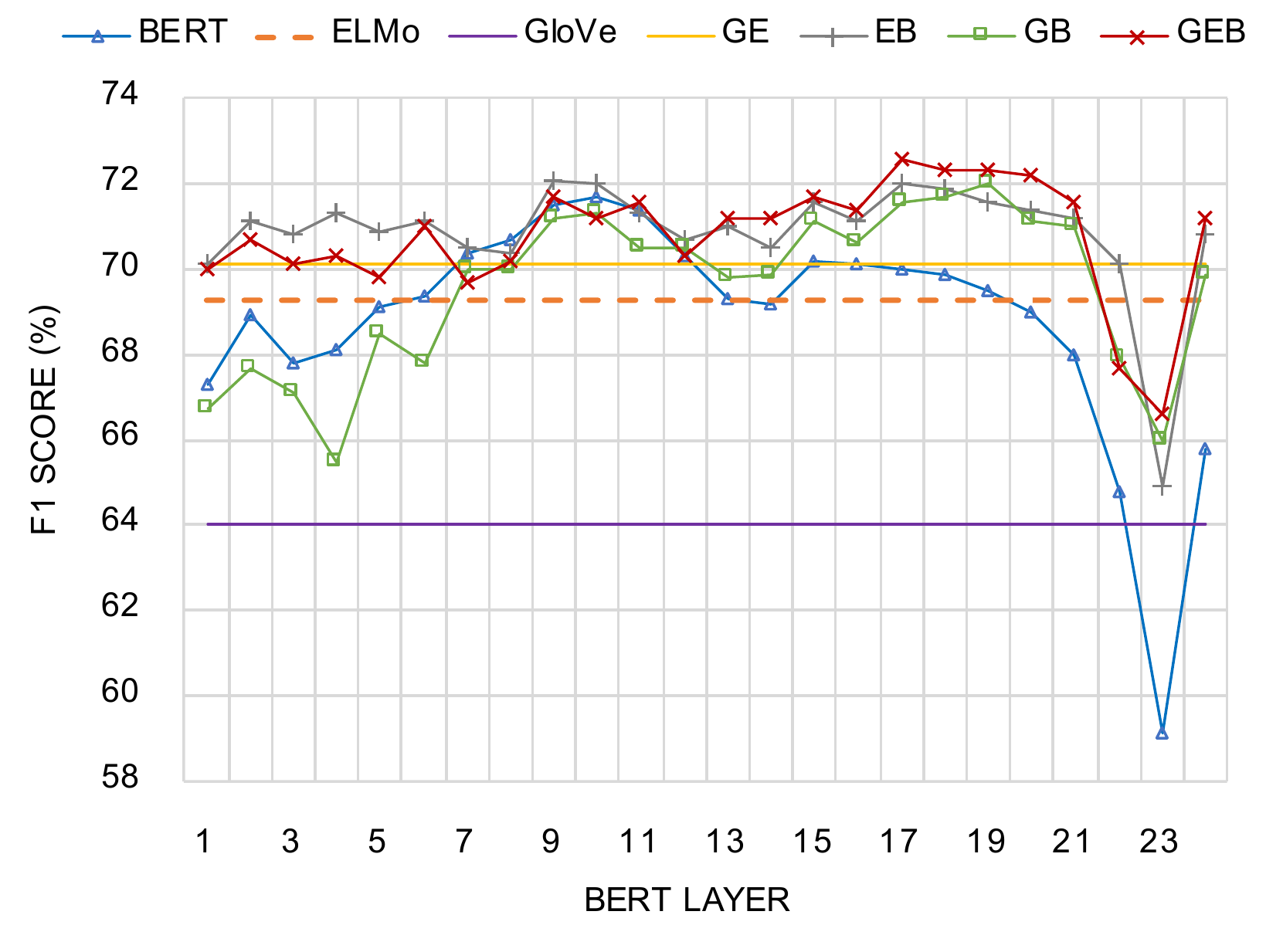}
    \caption{Model performance on VUA-all-PoS development set given by different word embedding features. BERT layer 1 is close to the input side, while layer 24 is close to the output side.}
    \label{fg:valid}
\end{figure}
\end{center}

\subsection{Different BERT layer analysis}
\label{sect: different BERT layer analysis}

In this section, we explore which BERT layers are more effective for the metaphor identification task. We focus on BERT rather than ELMo because previous works have already discovered that the most transferable layer of ELMo is the first layer as tested across a number of semantic and syntactic tasks~\cite{liu2019linguistic,schuster2019cross}. 

Figure~\ref{fg:valid} shows that both low BERT
layers$_{1:4}$ and top layers$_{21:24}$ yield weak results (blue line), while the middle layers$_{9:11}$ are strong. This is somewhat contradictory to the finding of \citet{devlin2019bert}, where they proposed to use the concatenation of the last four feature-based BERT layers$_{21:24}$ as the input features for the NER task. This implies that which BERT layer works best as a contextualised embedding is task dependent. 
Although ELMo (orange dash line) significantly outperforms GloVe (purple line), the combination of them (GE, yellow line) still yields better performance. Such a pattern can also be observed in GB$_{13:24}$ combinations (green line) compared with BERT layer$_{13:24}$ (blue line) and GloVe. Finally, GEB$_{17:20}$ (red line) outperform other word embeddings and their combinations across BERT layers, where GEB$_{17}$ is the best.

\subsubsection{Probing BERT layer effectiveness via word similarity}
\label{sect:Differentiating}

We hypothesise that the effectiveness of different contextualised embedding layers for metaphor detection can be probed by measuring how well the embedding can distinguish metaphors and literals that take the same word form. The underlying intuition is that  metaphoric meanings are strongly context dependent, and hence the task provides a natural challenging setting for testing the effectiveness of contextualised embeddings for the task. 

We conducted the experiment based on the VUA-all-PoS training set\footnote{The biggest one in our experimental datasets in terms of size.}, which contains 1,516 unique words that have both metaphoric and literal labels. For each of those words, we constructed a metaphor-literal word-pair set. For instance, the word-pair corresponding to ``wash'' contains  a literal ``wash'' from ``I'm supposed to wash up aren't I'' (\texttt{VUA\_ID: kbw-fragment04-2697}) and a metaphorical ``\textit{wash}'' from ``Do we really \textit{wash} down a good meal with claret?'' (\texttt{VUA\_ID: a3c-fragment05-233}). In the case where a word associated with a particular label appears in multiple sentences, we randomly sample one sentence for that label for constructing the word-pair. The total number of word-pairs is therefore 1,516.  

\begin{center}
\begin{figure}[t!]
    \centering
    \includegraphics[scale=0.44]{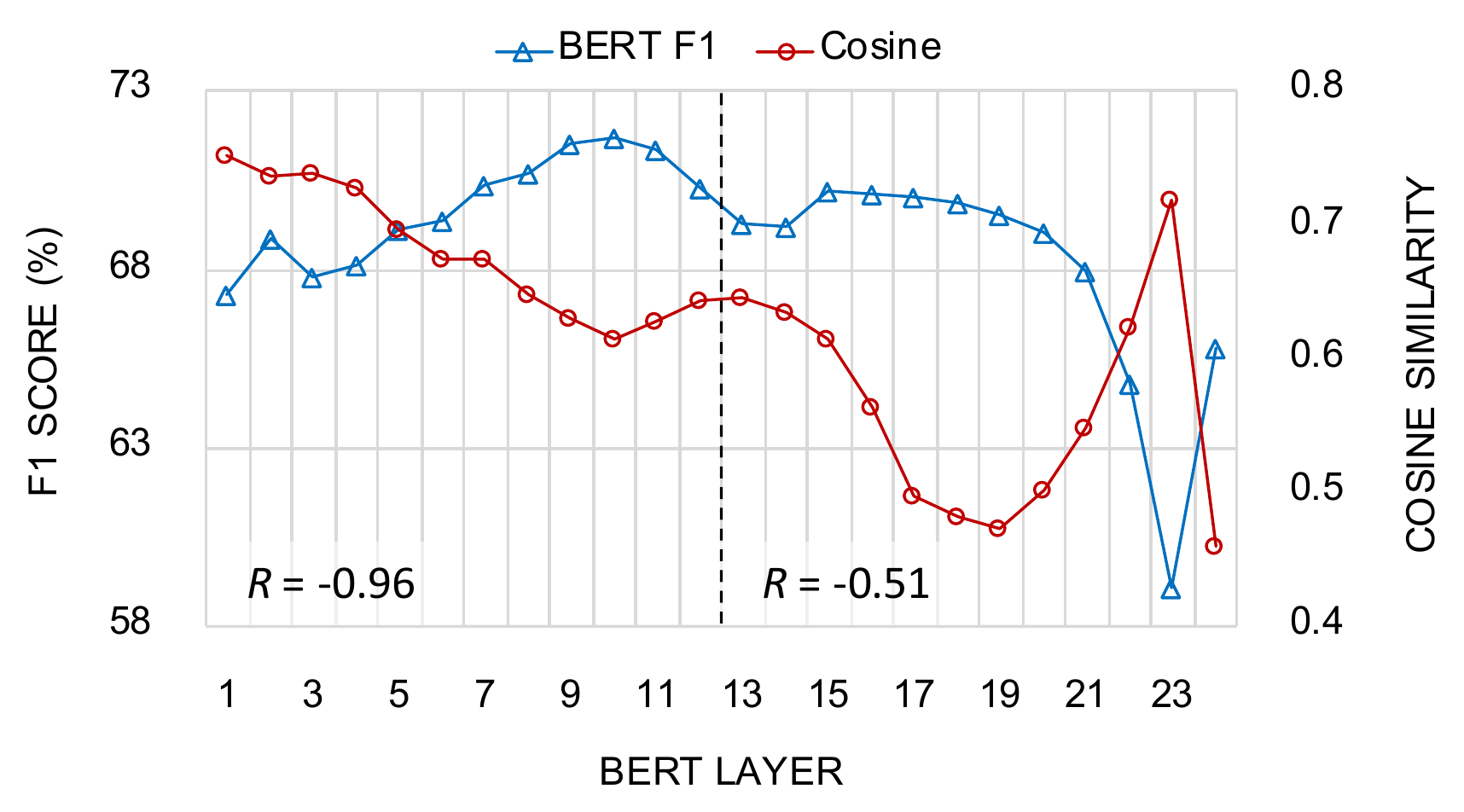}
    \caption{Correlation between model performance on VUA-all-PoS development set and BERT layer effectiveness on the training set. BERT layer effectiveness is represented by the average cosine similarity between metaphor-literal word-pairs (lower is better).}
    \label{fg: corr_bert_cos}
\end{figure}
\end{center}
\vspace{-2em}

To probe the effectiveness of different BERT layers, we compute the average cosine similarity between the words for all metaphor-literal word-pairs by taking the embedding from each layer as input. A lower average cosine similarity indicates a BERT layer is more effective in distinguishing metaphors from literals. The result is shown in the red line in Figure~\ref{fg: corr_bert_cos}. For comparison, we also show the metaphor identification performance given by a one-channel CNN+BiLSTM model using different BERT layer as input (blue line in Figure~\ref{fg: corr_bert_cos}). 

It can be observed from Figure~\ref{fg: corr_bert_cos} that there exists a negative correlation between the model performance in metaphor identification and how well the BERT embeddings can distinguish metaphors from literals, i.e., in general the lower average cosine similarity, the better model performance. In particular, the correlation is much stronger for the low level BERT layers (1:12; Pearson's $R=-0.96$) than the high level layers (13:24; $R=-0.51$). It can also be observed that the bottom BERT layers (1:4) mostly encoding general features are more effective than the top layers (21:24) which encode LM task-specific features. Thus, layer 24 which is the closest to LM output during BERT pre-training procedure does not yield the highest performance on metaphor identification, although layer 24 has the lowest cosine similarity. The middle layers (9:11), which capture both general and task-specific features, seem to work best with layer 10 giving the best performance. Such an observation is similar to the findings of \citet{liu2019linguistic}, who argued that the difference in the transferability across contextualiser layers on downstream tasks is due to the trade-off between embedded general and task-specific features. 




\begin{center}
\begin{figure}[t!]
    \centering
    \includegraphics[scale=0.45]{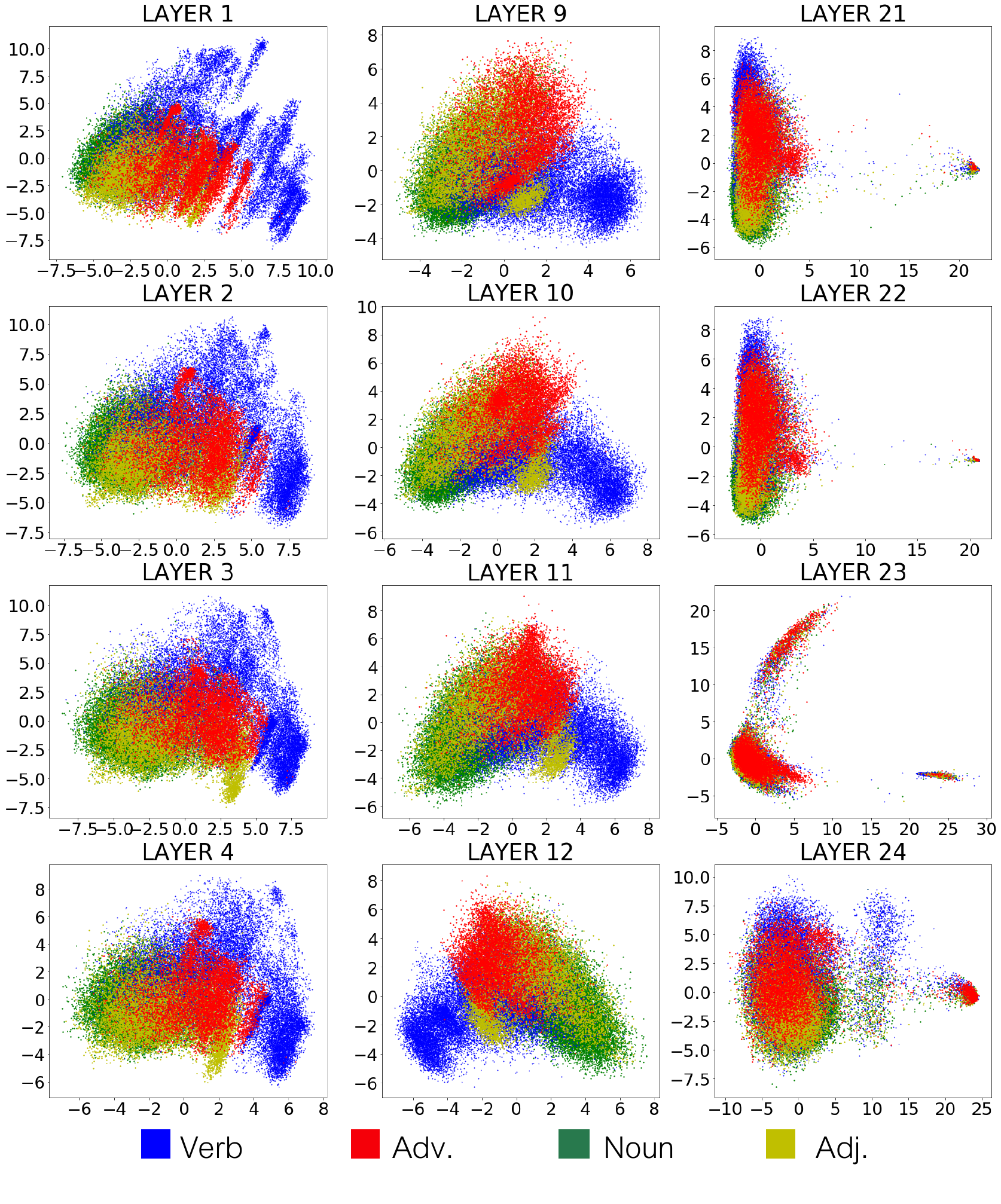}
    \caption{PCA-based 2D visualisations of BERT on open class words of the VUA-all-PoS training set.}
    \label{fg: pos_visualisation}
\end{figure}
\end{center}
\vspace{-2em}


\subsubsection{POS} \label{sect:POS}

We further visualise the distribution of word embeddings across different BERT layers to analyse how syntactic information is captured. 
For each BERT layer, we extract the embeddings for the open class words in the VUA-all-PoS training set, and then project the embeddings to a 2D space using Principal Component Analysis (PCA) \citep{jolliffe2011principal}. 
We show in Figure~\ref{fg: pos_visualisation} the distributions of the most representative BERT layers, i.e., 1:4, 9:12, and 21:24. Each point corresponds to a word and the colour indicates the word class. 

It can be observed from Figure~\ref{fg: pos_visualisation} that the distribution of the top layers (21:24) is highly concentrated in a relatively small region (layer 23 in particular), where words of different PoS classes are heavily overlapped with each other. In contrast, the word distributions of layers 1:4 and 9:12 exhibit a distinctive pattern, where words are more spread out and there are clear boundaries between words of different classes, showing that syntactic features like PoS are captured in those layers. These phenomena also explain why the effectiveness of explicitly incorporating PoS features is small, i.e., recalling the marginal gain of PoS+Abst+GEB$_{17}$ over GEB$_{17}$ discussed in $\S$~\ref{sect: overall performance}. 
To sum up, which BERT layer is more transferable for metaphor identification appears to depend on the trade-off between the general and task-specific features captured by the layer. 

\begin{table*}[tb]
\small
\centering
\begin{tabular}{>{\centering\arraybackslash}p{1.3cm}|>{\centering\arraybackslash}p{0.42cm}>{\centering\arraybackslash}p{0.42cm}>{\centering\arraybackslash}p{0.42cm}>{\centering\arraybackslash}p{0.42cm}|>{\centering\arraybackslash}p{0.42cm}>{\centering\arraybackslash}p{0.42cm}>{\centering\arraybackslash}p{0.42cm}>{\centering\arraybackslash}p{0.42cm}|>{\centering\arraybackslash}p{0.42cm}>{\centering\arraybackslash}p{0.42cm}>{\centering\arraybackslash}p{0.42cm}>{\centering\arraybackslash}p{0.42cm}|>{\centering\arraybackslash}p{0.42cm}>{\centering\arraybackslash}p{0.42cm}>{\centering\arraybackslash}p{0.42cm}>{\centering\arraybackslash}p{0.42cm}}
\Xhline{2.5\arrayrulewidth}

\bf Feature & \bf P & \bf R & \bf F1 & \bf Acc & \bf P & \bf R & \bf F1 & \bf Acc & \bf P & \bf R & \bf F1 & \bf Acc & \bf P & \bf R & \bf F1 & \bf Acc\\
\hline
 & \multicolumn{4}{c|}{\bf Academic} & \multicolumn{4}{c|}{\bf Conversation} & \multicolumn{4}{c|}{\bf Fiction} & \multicolumn{4}{c}{\bf News} \\
\cline{2-17}
GloVe   & 65.2 & 67.5 & 66.3 & 92.1      & 58.4 & 62.6 & 60.4 & 94.1      & 60.1 & 55.6 & 57.8 & 91.4      & 69.3 & 64.9 & 67.0 & 90.1 \\
ELMo        & 65.1 & 74.1 & 69.3 & 92.4      & 67.6 & 65.1 & 66.4 & 95.2      & 62.3 & 68.4 & 65.2 & 92.3      & 72.6 & 73.4 & 73.0 & 91.6 \\
BERT$_{17}$ & 67.3 & 71.7 & 69.4 & 92.7      & 70.9 & 63.0 & 67.7 & 95.7      & 70.3 & 65.9 & 68.1 & 93.5      & 74.0 & 71.1 & 72.6 & 91.6 \\ \cdashline{2-17}
GE          & 66.9 & 74.6 & 70.5 & 92.8      & 63.3 & 69.3 & 66.1 & 94.9      & 65.8 & 65.5 & 65.7 & 92.8      & 73.1 & 74.5 & 73.8 & 91.8 \\
GB$_{17}$   & 64.7 & 77.2 & 70.4 & 92.5      & 68.1 & 67.5 & 67.8 & 95.7      & 70.3 & 67.6 & 68.9 & 93.7      & 74.3 & 71.5 & 72.9 & 91.7  \\
EB$_{17}$   & 71.8 & 72.3 & 72.0 & 93.5      & 69.9 & 66.3 & 68.1 & 95.8      & 72.9 & 64.8 & 68.6 & 93.6      & 76.1 & 70.5 & 73.2 & 91.9 \\ \cdashline{2-17}
GEB$_{17}$  & 72.7 & 72.0 & \bf 72.3 & 93.8      & 74.0 & 64.9 & \bf 69.1 & 95.9      & 75.9 & 67.1 & \bf 71.2 & 94.3      & 77.7 & 71.4 & \bf 74.4 & 92.4 \\
\hline
\hline
 & \multicolumn{4}{c|}{\bf Verb} & \multicolumn{4}{c|}{\bf Adjective} & \multicolumn{4}{c|}{\bf Noun} & \multicolumn{4}{c}{\bf Adverb} \\
\cline{2-17}
GloVe       & 60.2 & 57.2 & 58.7 & 84.9      & 54.9 & 42.2 & 47.7 & 90.1      & 59.1 & 50.5 & 54.5 & 88.6      & 49.4 & 49.4 & 49.4 & 93.0 \\
ELMo        & 62.7 & 70.3 & 66.3 & 86.6      & 46.7 & 54.9 & 50.5 & 88.5      & 61.5 & 58.6 & 60.0 & 89.5      & 57.6 & 51.9 & 54.6 & 94.0 \\
BERT$_{17}$ & 63.3 & 72.2 & 67.5 & 89.9      & 54.7 & 49.1 & 51.8 & 90.2      & 66.8 & 51.7 & 58.3 & 90.0      & 66.7 & 45.5 & 54.1 & 94.7 \\ \cdashline{2-17}
GE          & 62.4 & 68.9 & 65.5 & 86.4      & 56.9 & 58.7 & \bf 57.8 & 90.8      & 62.4 & 59.9 & 61.1 & 90.1      & 53.7 & 56.5 & 55.1 & 93.6 \\
GB$_{17}$   & 64.7 & 69.1 & 66.8 & 87.1      & 58.4 & 53.8 & 56.0 & 90.9      & 65.0 & 57.7 & 61.1 & 90.1      & 61.3 & 49.4 & 54.7 & 94.3 \\
EB$_{17}$   & 66.9 & 69.0 & 67.9 & 87.8      & 53.7 & 53.2 & 53.4 & 90.1      & 73.4 & 49.5 & 59.1 & 90.8      & 63.3 & 49.4 & 55.5 & 94.5 \\ \cdashline{2-17}
GEB$_{17}$  & 71.6 & 67.4 & \bf 69.4 & 88.9      & 62.8 & 53.5 & \bf 57.8 & 91.6      & 69.9 & 54.5 & \bf 61.3 & 90.7      & 69.1 & 49.4 & \bf 57.6 & 95.0 \\
\Xhline{2.5\arrayrulewidth}
\end{tabular}
\caption{The utilities of different word embeddings and combinations on different genres and PoS of VUA-all-PoS development set.}
\label{tb: breakdown}
\end{table*}

\subsection{Effectiveness of combining multiple word embeddings} \label{sect: compensations analysis} 


In order to analyse the semantic and syntactic complementarity between different pre-trained word embeddings, we analyse the performance of each feature based on different genres and PoS. We examine their performance on VUA-all-PoS development set because its sentences were labelled with multiple genre categories, e.g., academic text, fiction, news and conversation. Compared with MOH-X and TroFi whose annotated target words focus on verbs, VUA provides metaphor annotation to all word classes. Here, we examine the model performance on open class words (i.e., verbs, adjectives, adverbs, and nouns) because they consist of the majority of metaphors in the VUA dataset. 

It can be seen from Table~\ref{tb: breakdown} that ELMo slightly outperforms BERT$_{17}$ on news (0.4\%) where BERT$_{17}$ gives better performance on the other three genres with the biggest difference on the fiction category (2.9\%), which could be explained by the fact that ELMo was trained on the WMT 2011 News Crawl data whereas BERT was trained on Wikipedia and BookCorpus.
The biggest advantage of BERT$_{17}$ against ELMo appearing on fiction also supports such a hypotheses, i.e., the BookCorpus provides good training resources for BERT in the book domain, and hence BERT can better handle the metaphors in fiction. These observations show that that our model can incorporate the strengths of different word embeddings in different domains, so that we do not have to re-train a supermodel on a very large multi-genre corpus. 

It can  be observed that contextualised embeddings i.e., ELMo and BERT outperform GloVe by large margins of 5.6\% and 6.6\% on average, respectively. This is somewhat unsurprising as the task of metaphor identification is strongly context dependent. There is also a clear pattern that the combination of two embeddings works better than  individual embeddings, and the combination of three embeddings works best. EB$_{17}$, the best two-embedding combination, outperforms the best single embedding BERT by an average of 1.0\% in F1. Using the three embedding combination further boosts the model performance where a gain of 2.3\% over the best performing single embedding BERT is achieved. This clearly shows the benefit of using multiple embeddings for metaphor identification on different genres. 



For PoS, similar patterns to the genre analysis can be observed, where incorporating two embeddings gives better performance than each individual embedding and that incorporating three embeddings works best. For adjectives in particular, there is a big improvement of GE over ELMo (7.3\%) and GB$_{17}$ over BERT$_{17}$ (4.2\%).  
Overall, the combination of different word embeddings presents the strongest performance across different genres and PoS, which is mainly due to the difference in training algorithms and corpora between the three selected word embeddings.

\subsection{Bad combination analysis}
\label{sect: bad combination analysis}

\begin{center}
\begin{figure}[t!]
    \centering
    \includegraphics[scale=0.45]{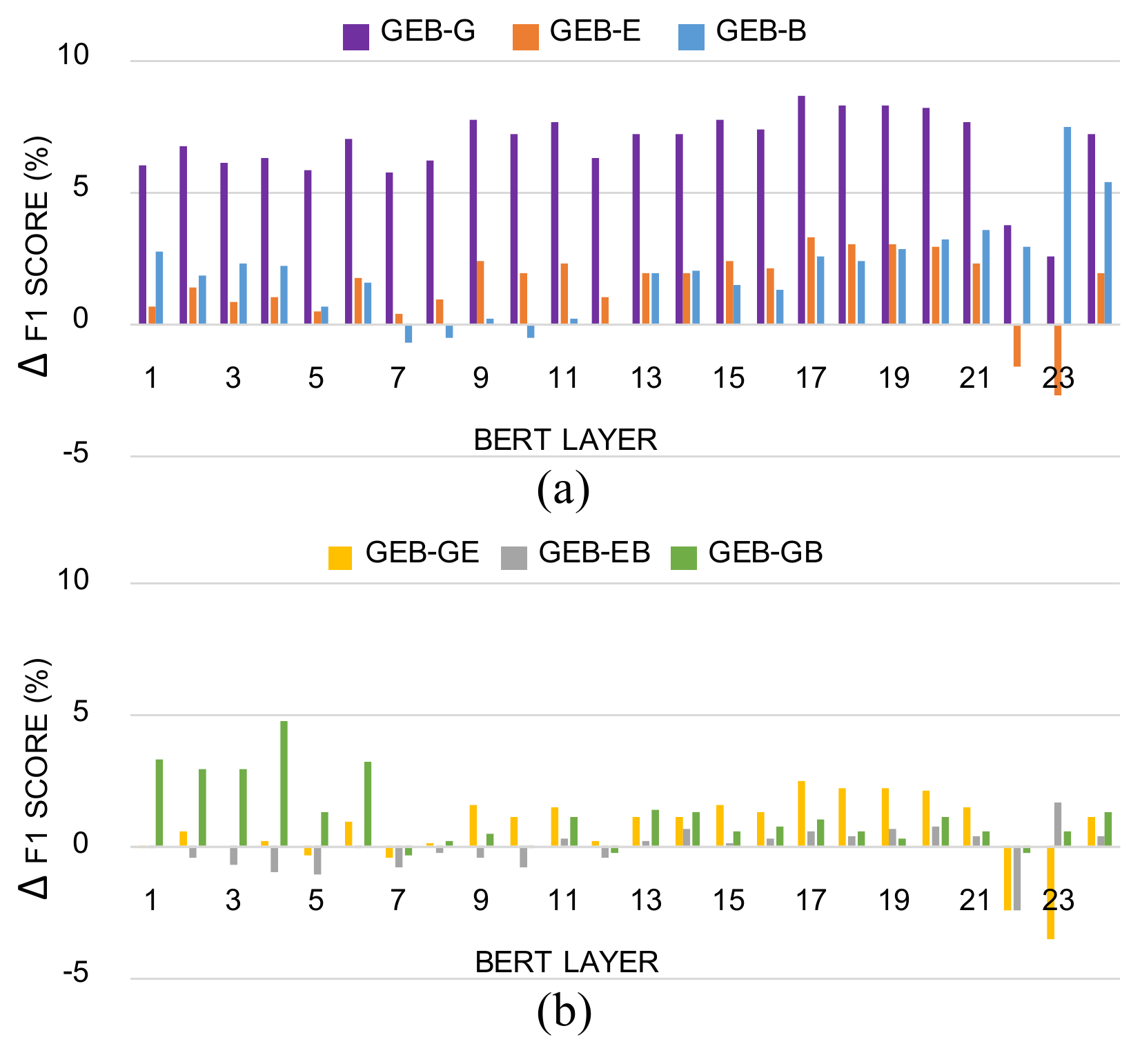}
    \caption{Model performance differences between GEB and other word embedding setups on VUA-all-PoS development set.}
    \label{fg:ebdifference}
\end{figure}
\end{center}

We demonstrate the difference in model performance ($\Delta$ F1), given by different word embedding setups in Figure \ref{fg:ebdifference}. As seen, although GEB outperforms most of the single word embedding based models in Figure \ref{fg:ebdifference}a in terms of F1 score, we also observe that EB surpasses GEB before BERT layer 11 in Figure \ref{fg:ebdifference}b. An intuitive question is what makes bad combinations across BERT layers. 
Apart from the trade-off between general and task-specific features (see $\S$ \ref{sect: different BERT layer analysis}), we try to explain the bad combinations from spatial relations between ELMo and BERT layers here.

\vspace{-1em}
\begin{center}
\begin{figure}[t!]
    \centering
    \includegraphics[scale=0.45]{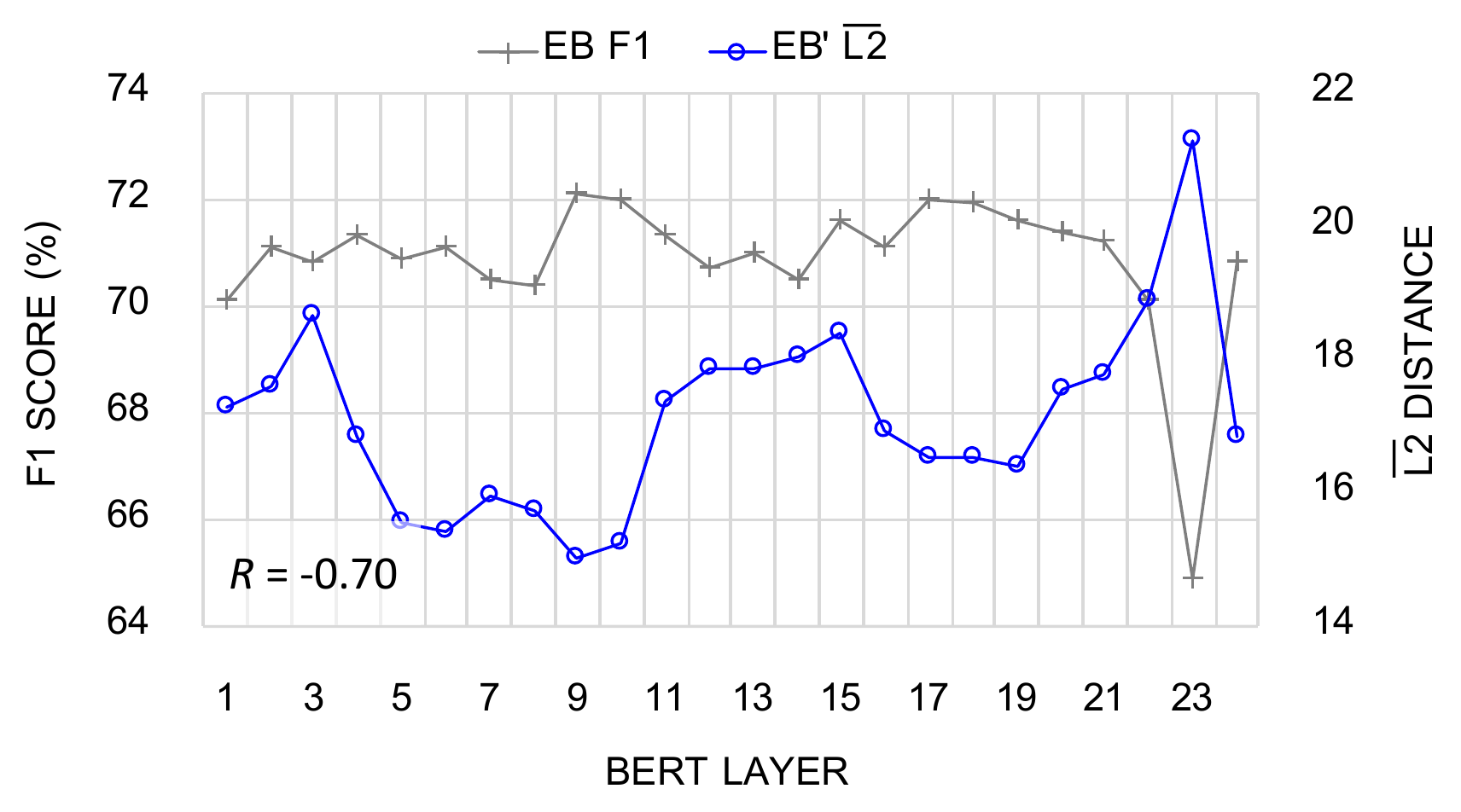}
    \caption{Correlation between EB model performance on VUA-all-PoS development set and EB$'$ average $L2$ distance on the training set.}
    \label{fg:correlation}
\end{figure}
\end{center}
\vspace{-2em}

We analyse the combinations of ELMo and different BERT layers, because they are aligned in dimensionality. BERT vectors are rotated to ELMo space with an orthogonal mapping approach \cite{xing2015normalized,conneau2017word}.
\begin{equation}
    W^{*}=\operatorname*{argmin}_{W \in O_{d}(\mathbb{R})} \left \| WB-E \right \|_{F} = UV^{\top}
\end{equation}
with $U \Sigma V^{\top}= {\rm SVD}(EB^{\top})$, where $O_{d}(\mathbb{R})$ is $1024 \times 1024$ dimension matrices of real numbers. $B$ and $E$ are BERT and ELMo embedding matrices of the VUA-all-PoS training set. $W$ is the learnt linear mapping matrix. SVD is Singular Value Decomposition. The rotated BERT ($B'$) is given by $WB$. We compute the average $L2$ distance ($\overline{L2}$) between $E$ and each layer of $B'$ with 
\begin{equation}
\label{eq:l2}
    \overline{L2}_{EB'} = \frac{1}{n}\sum_{i=1}^{n}\sqrt{\sum_{j=1}^{d}(E_{i,j}-B'_{i,j})^2}, 
\end{equation}
where $n$ is the total number of tokens in the training set ($n=116,622$); $i$ is the position of a token; $j$ is the position of an element of vectors in $E$ and $B'$. We compare the $\overline{L2}$ with F1 scores on VUA-all-PoS development set. 

As seen in Figure~\ref{fg:correlation}, there is a negative correlation ($R=-0.70$) between $\overline{L2}$ and model performance\footnote{Using non-rotated BERT instead of $B'$ in Eq.~\ref{eq:l2} also presents a negative correlation with the model performance, whereas the correlation is weaker ($R=-0.53$).}. 
The selected parallel word embeddings are distant from each other in vector space, meaning that the distribution patterns of the two embeddings are likely very different. A model may struggle to combine distant word embeddings,
as they present less consistency in representing word semantics and syntax.

\section{Conclusion}
We propose a model that can incorporate advanced word embedding and linguistic features, achieving state-of-the-art performance on the sequential metaphor identification task. We also examine different combinations of GloVe, ELMo and different BERT layers. Our experiments show that the output-side BERT layers do not distinguish metaphors and literals well, which is different to the finding of \citet{devlin2019bert} where the last four layers of BERT yielded strong performance on the NER task. In contrast, the middle layers of BERT perform best for sequential metaphor identification. We also find that different word embeddings can complement each other, because they may embed complementary semantics and syntax, due to training on different corpora with different algorithms. This offers opportunities for leveraging multiple existing pre-trained word embeddings for improving a variety of downstream tasks, which we would like to explore in the future.



\bibliography{tacl2018}
\bibliographystyle{acl_natbib}


\end{document}